\documentclass{article} % For LaTeX2e
\usepackage{continual_learning,times}

\usepackage{amsmath,amsfonts,bm}

\def\eqref#1{equation~\ref{#1}}
\def\1{\bm{1}}

\DeclareMathAlphabet{\mathsfit}{\encodingdefault}{\sfdefault}{m}{sl}
\SetMathAlphabet{\mathsfit}{bold}{\encodingdefault}{\sfdefault}{bx}{n}

\usepackage{hyperref}
\usepackage{url}

\usepackage[utf8]{inputenc} % allow utf-8 input
\usepackage[T1]{fontenc}    % use 8-bit T1 fonts
\usepackage{booktabs}       % professional-quality tables
\usepackage{amsfonts}       % blackboard math symbols
\usepackage{nicefrac}       % compact symbols for 1/2, etc.
\usepackage{microtype}      % microtypography

\usepackage{mathtools}
\usepackage{algorithm, algpseudocode}
\usepackage{booktabs}
\usepackage[toc,page]{appendix}

\title{Better Knowledge Retention through Metric Learning}

\author{%
Ke Li\thanks{Denotes equal contribution.}\\
Simon Fraser University, Google \&\\
Institute for Advanced Study\\
\texttt{keli@sfu.ca} \\
\And
Shichong Peng$^{\ast}$ \\
Simon Fraser University \&\\
University of California San Diego\\
\texttt{shichong\_peng@sfu.ca}\\
\And
Kailas Vodrahalli$^{\ast}$ \\
Stanford University\\
\texttt{kailasv@stanford.edu}\\
\And
\hskip 110pt Jitendra Malik \\
\hskip 110pt University of California, Berkeley\\
\hskip 110pt \texttt{malik@eecs.berkeley.edu}
}

\iclrfinalcopy
\begin{document}

\maketitle

\begin{abstract}
In continual learning, new categories may be introduced over time, and an ideal learning system should perform well on both the original categories and the new categories. While deep neural nets have achieved resounding success in the classical supervised setting, they are known to forget about knowledge acquired in prior episodes of learning if the examples encountered in the current episode of learning are drastically different from those encountered in prior episodes. In this paper, we propose a new method that can both leverage the expressive power of deep neural nets and is resilient to forgetting when new categories are introduced. We found the proposed method can reduce forgetting by $2.3\times$ to $6.9\times$ on CIFAR-10 compared to existing methods and by $1.8\times$ to $2.7\times$ on ImageNet compared to an oracle baseline.
\end{abstract}

\section{Introduction}
\label{sec:introduction}

Despite the multitude of successes of deep neural networks in recent years, one of the major unsolved issues is their difficulty in adapting to new tasks while retaining knowledge acquired from old ones. This problem is what continual learning aims to tackle. To successfully develop a continual learning method, one major limitation that must be overcome is the problem of \emph{catastrophic forgetting}~\citep{mccloskey1989catastrophic}, which describes the phenomenon that a machine learning model fine-tuned for a new task performs poorly on the old task it was originally trained on. 

Various continual learning methods make different assumptions about whether the task identities and boundaries are known at training time and test time~\citep{van2019three,hsu2018re,Zeno2018TaskAC}. In the setting of \emph{task learning}, task identities are known at both training and test time and the method is only asked to classify among the possible classes \emph{within} the given task. In \emph{domain learning}, the requirement for task identity to be known at test time is removed (with all others in place) and each task is assumed to have the same number of classes with similar semantics; in this case, the method is asked to classify among the possible classes within any given task. \emph{Class learning} differs from \emph{domain learning} in that each task may contain a different number of classes that may have non-overlapping semantics with the classes in other tasks, and the method is asked to classify among the possible classes across \emph{all} tasks. \emph{Discrete task-agnostic learning} further generalizes this by removing the assumption of known task identity at training time and replacing it with an assumption of known boundaries between different tasks at training time. In the most general setting, task boundaries between different tasks are also unknown, even at training time. This setting is known as \emph{continuous task-agnostic learning}, which is the setting we focus on in this paper. 

Existing continual learning methods can be divided into two broad categories: those that bias the parameters towards parameter values learned on old tasks, and those that expand the model size to accommodate new tasks. In this paper, we propose a new approach that is orthogonal to these categories. Our approach neither penalizes parameter changes nor expands model size. Our key intuition is that neural nets forget because parameters at all layers need to change to adapt to new tasks. Therefore, one way to improve retention is to keep parameter changes localized. Because upper-layer parameters depend on lower-layer parameters, upper-layer parameters must change when lower-layer parameters change; hence, lower-layer parameters must be kept relatively static to prevent parameter changes throughout the network. One way to achieve this would be to explicitly regularize parameter changes in the lower layers; this is less than ideal, however, because it would reduce the network's expressive power and therefore its capability to learn on new tasks. How do we achieve this without compromising on expressive power?

To arrive at a solution to this problem, we need to consider the underlying cause of parameter changes at \emph{all} layers when fine-tuning on new tasks. In a neural net, each layer on its own has quite limited expressive power and is simply a generalized linear model. As a result, if the features computed by the penultimate layer on training examples for the new task are not linearly separable, then the parameters in the layers up to the penultimate layer must change to successfully learn the new task. To avoid this, we can make the last layer more expressive and replace it with a nonlinear classifier. To this end, we propose replacing the softmax activation layer with a $k$-nearest neighbour classifier. We will show that this simple but powerful modification is surprisingly effective at reducing catastrophic forgetting.  

\section{Related Work}
\label{sec:related_work}

Our work offers a novel solution to the continual learning problem by leveraging ideas from metric learning.

\textbf{Continual learning} is the idea that machine learning models should be able to emulate humans' ability to learn new information while retaining previously learned skills or knowledge. This capability is generally relevant -- for example, in robotics it is desirable for robots to adapt to a given environment while retaining a general functionality across environments; another example from computer vision is the desired ability to update a classifier online with information on new classes while retaining the ability to detect old classes. While it is always possible to accomplish this goal of continual learning through retraining a model on the combined old and new datasets (e.g., as in \citep{caruana1997multitask}), this approach is computationally expensive or infeasible due to the inability to store old data. One commonly used shortcut is to instantiate a model trained on the old dataSet And then \textit{fine-tune} that model on the new data (e.g., as in \citep{hinton2006reducing, girshick2014rich}); however, this method has the side effect of (significantly) decreased performance on the old data. 

% new version
Recent interest in continual learning has resulted in many proposed methods for continual learning with deep neural networks. They can be grouped into two broad categories: those that dynamically expands the model architecture, and those that constrains the parameter values to be close to those obtained after training on previous tasks. 

An example of a method in the former category is \citep{rusu2016progressive}, which creates a new network for each new task, with connections between the networks (from old task networks to new ones), thus ensuring the network weights for the old network are preserved while enabling adaptation to the new task. Another method~\citep{Masse2018AlleviatingCF} randomly zeros a subset of hidden units for each task, which can be viewed as a way to implicitly expand the model architecture, since the hidden units that are zeroed are effectively removed from the architecture. 

In the latter category are two subtypes of methods: those based on regularization and those based on data replay. Regularization-based methods are all based on the idea of protecting important parameters and introduce different regularizers to penalize changing the values of these parameters from what they were after training on previous tasks. Different methods differ in the way they determine importance of each parameter. Examples of methods of this flavour include elastic weight consolidation (EWC)~\citep{kirkpatrick2017overcoming}, online EWC~\citep{schwarz2018progress} and synaptic intelligence (SI)~\citep{Zenke2017ContinualLT}. 

Whereas regularization-based methods can be viewed as constraining the parameter values explicitly, data replay-based methods do so implicitly. Generally speaking, data replay works by generating simulated training data from the model trained on previous tasks and using it to augment the training data used to train on the new task. For example, Learning without Forgetting (LwF)~\citep{li2018learning} performs distillation~\citep{Hinton2015DistillingTK} on the training data for the new task, or in other words, takes simulated labels to be the predictions of the model trained on previous tasks on the input training data for the new task. Deep Generative Replay (DGR)~\citep{Shin2017ContinualLW} uses a deep generative model to generate simulated input data and labels them with the model trained on previous tasks. Other related methods~\citep{rebuffi2017icarl,Wu2018IncrementalCL, Venkatesan2017ASF, Ven2018GenerativeRW} use training data from previous tasks, either directly or to train a deep generative model, the samples from which are used to augment the current training set. For a more comprehensive review of this area, we refer readers to \citep{parisi2019continual}.

Note that all these methods at least require task boundaries to be known. For example, methods that dynamic expand the model architecture need to know when to expand the architecture, regularization-based methods need to know the point in time at which the parameter values are used to constrain future parameter updates, and data replay-based methods need to know when to generate simulated training data. Some methods, like \citep{Masse2018AlleviatingCF}, require stronger assumptions, namely that task identities be known as well. In contrast, we consider the setting where neither task identities nor task boundaries are known. 

\textbf{Metric learning} involves learning a (pseudo-) distance metric. This can be combined with $k$-nearest neighbours ($k$-NN) classifier to yield classification predictions. Standard $k$-NN works by finding the closest $k$ training data points to a test data point and classifying the test data point as the most common label among the training data. The distance metric used for determining the ``closest'' $k$ points is commonly the Euclidean distance. As proposed by \citet{weinberger2006distance}, it is possible and desirable to learn a Mahalanobis distance metric that improves the accuracy of $k$-NN. The loss function attempts to enforce clustering of labels with large margins between clusters.

More recent work like \citep{schroff2015facenet} has used similar ideas in the context of deep neural networks where the distance between two data points is defined as the Euclidean distance between their embeddings computed by the network. Most methods work by pulling examples whose labels are the same as that of the current training example closer to the training example and pushing examples whose labels are different away. The challenge becomes designing a loss function that is conducive to efficient training of the network. Various loss functions like those used in \citep{schroff2015facenet, tamuz2011adaptively, van2012stochastic} have been proposed, to varying degrees of success on varying datasets.

\section{Method}
\label{sec:method}

\subsection{Model}

The key observation we make is that in order for a neural net to learn on non-linearly separable data, the parameters in the layers below the last layer must change. So, if the training data for the new task is not linearly separable from the training data for the previous tasks, the parameters below the last layer will be changed when training on the new task. Because the parameters in the layer below the last layer are shared across all classes, the classes seen during previous tasks must depend on the parameters below the last layer. And so changes in those parameters would result in a performance drop on these classes, leading to catastrophic forgetting. 

To avoid catastrophic forgetting, we must make sure the parameters below the last layer will not change significantly even as the model is trained on new tasks. To this end, we must make the last layer more expressive than a linear classifier. Therefore, we replace it with a nonlinear classifier, namely a $k$-nearest neighbour classifier using a learned Mahalanobis distance metric, which is equivalent to replacing the softmax activation layer in a neural net with a $k$-nearest neighbour classifier using a Euclidean distance metric. 

\subsection{Learning}

\subsubsection{Loss Function}

We would like to design the loss that encourages the predictions of the $k$-nearest neighbour classifier on the feature activations of the last layer to be accurate. To this end, for every training example, we would like the nearby training examples to be in the same class and the training examples that are in other classes to be far away. In practice, we found it suffices to enforce this for only a small subset of training examples, which we will refer to as \emph{anchors}. The training examples that are in the same class as an anchor is known as \emph{positive examples}, and those that are in different classes are known as \emph{negative examples}. In practice, we choose one anchor for each class.

\begin{figure}[H]
    \centering
    \includegraphics[width=5in]{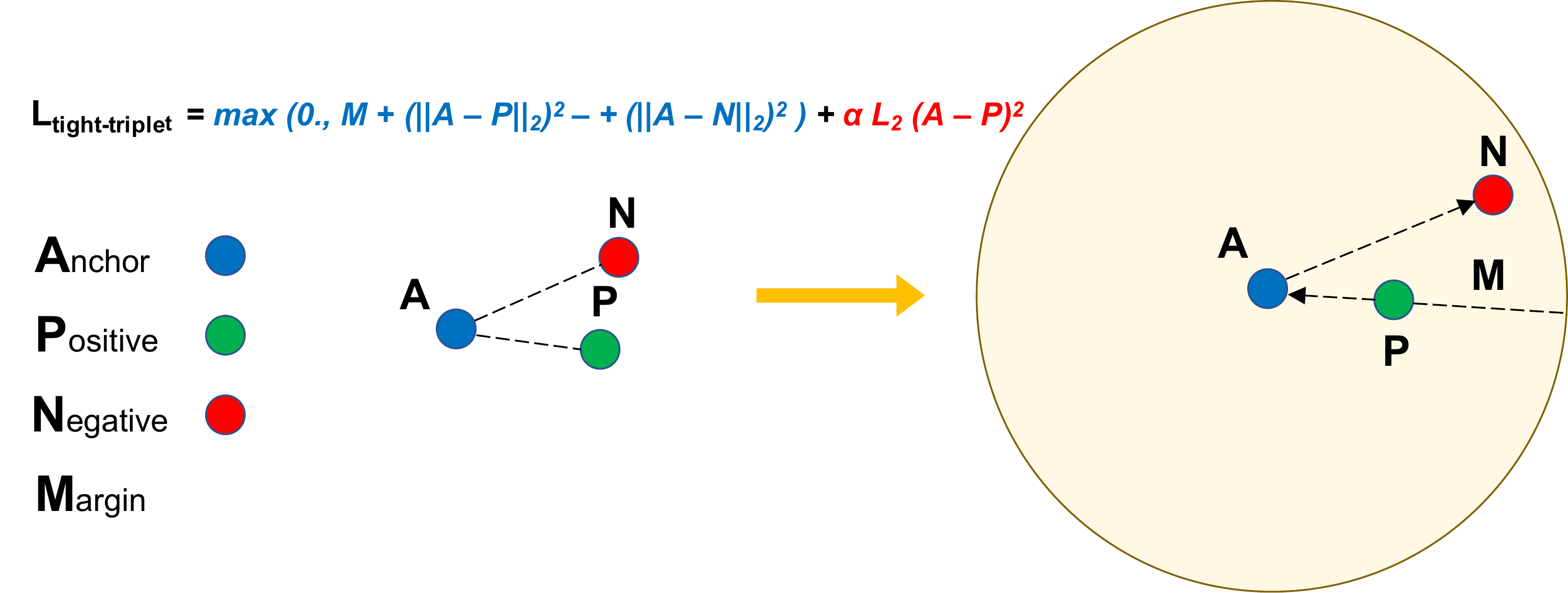}
    \caption{Graphical depiction of the loss function in Eq.~\ref{eq:triplet_loss}.}
    \label{fig:loss_function}
\end{figure}

The loss function we use is a modified form of the triplet loss~\citet{schroff2015facenet}, with an added term to encourage the examples of the same class to form a cluster:
\begin{equation}
\begin{aligned}
    L = \sum_{i=1}^{N} \max \left( \Vert\tilde{\Phi}(x_i^a) - \tilde{\Phi}(x_i^p)\Vert_2^2 \right. - \left.\Vert\tilde{\Phi}(x_i^a) - \tilde{\Phi}(x_i^n)\Vert_2^2 + M, 0 \right) + \alpha \sum_{i=1}^{N} ||\tilde{\Phi}(x_i^a) - \tilde{\Phi}(x_i^p)||_2^2 \label{eq:triplet_loss}
\end{aligned}
\end{equation}
where $\tilde{\Phi}$ denotes the model, i.e.: a function that maps the input to embedding, $x_i^a$ denotes an anchor, $x_i^p$ denotes a positive example and $x_i^n$ denotes a negative example, $M$ is a hyperparameter representing the desired minimum margin of separation between the positive and negative examples and $\alpha$ is a hyperparameter for balancing two terms in the loss.

\subsubsection{Training Procedure}

\begin{algorithm}
    \caption{Training Procedure}
    \label{alg:training_procedure}
    \begin{algorithmic}
        \State Sample set $X_a$ of $k_a$ anchors
        \For{$l = 1, \ldots$}
            \State Sample subset $X_e$ of $k_e$ training examples
            \For{$x_{ai} \in X_a$}
                \State $X_p \gets $ set of positive examples of $x_{ai}$ in $X_e$
                \State $X_n \gets $ set of negative examples of $x_{ai}$ in $X_e$
                \State $D_{ai,pj} \gets ||\tilde{\Phi}(x_{ai})-\tilde{\Phi}(x_{pj})||_2^2 \qquad\forall x_{ai} \in X_a, x_{pj} \in X_p$
                \State $D_{ai,nk} \gets ||\tilde{\Phi}(x_{ai})-\tilde{\Phi}(x_{nk})||_2^2 \qquad\forall x_{ai} \in X_a, x_{nk} \in X_n$
            \EndFor
            \For{$x_{ai} \in X_a$} 
                \State $T_{x_{ai}} \gets \{(x_{ai}, x_{pj}, x_{nk}) : x_{pj} \in X_p, x_{nk} \in X_n,~ D_{ai,pj} < D_{ai,nk}\}$ 
                \State $\tilde{T}_{x_{ai}} := \{(t_a, t_p, t_n) \in T_{x_{ai}} : t_n\mbox{ is among the }d\mbox{ closest negative examples to }t_a\mbox{ in }T_{x_{ai}}\}$
                \State Train model $\tilde{\Phi}$ on $\tilde{T}_{x_{ai}}$ for $m$ steps
            \EndFor
        
        \EndFor
         
    \end{algorithmic}
\end{algorithm}

\begin{table*}[t]
\centering
\footnotesize
\begin{tabular}{lccc}
\toprule 
Method & Set A: \textit{Original} & Set A: \textit{After Seeing Set B} & Set B\\
&&(Difference from \textit{Original})&\\
\midrule
  Our Method & 97.5\% & 95.1\%\ (-2.4\%) & 95.6\% \\
\midrule
\multicolumn{4}{l}{\emph{Very Forgetful \& Adaptive:}}\\
  Fine-tuning with dropout (ratio: 0.5) & 98.62\% & 0.0\%\ (-98.62\%) & 98.62\% \\
  Elastic weight consolidation (EWC) & 98.62\% & 0.0\%\ (-98.62\%) & 98.46\% \\
  Online EWC & 98.62\% & 0.0\%\ (-98.62\%) & 98.46\% \\
  Synaptic intelligence (SI) & 98.62\% & 0.0\%\ (-98.62\%) & 98.13\% \\
\midrule
\multicolumn{4}{l}{\emph{Forgetful \& Adaptive:}}\\
  Deep Generative Replay (DGR) & 98.78\% & 89.32\%\ (-9.46\%) & 98.40\% \\
  DGR with distillation & 98.78\% & 91.70\%\ (-7.08\%) & 98.05\% \\
  Replay-trough-Feedback (RtF) & 98.96\% & 93.22\%\ (-5.74\%) & 98.62\% \\
\midrule
\multicolumn{4}{l}{\emph{Not Forgetful \& Not Adaptive:}}\\
  Learning without Forgetting (LwF) & 98.62\% & 98.07\%\ (-0.55\%) & 6.84\% \\
\bottomrule
\end{tabular}
\caption{Performance of the proposed method compared to eight other continual learning methods on MNIST. SGD with dropout, EWC, online EWC and SI suffer from catastrophic forgetting on Set A after seeing Set B, but were able to adapt well to the new task (Set B). DGR, DGR + distillation and RtF forget more on Set A compared to our method. LwF retains knowledge learned on Set A; however, it fails to adapt to Set B. We also note that all baseline methods assume knowledge of task boundaries, whereas the proposed method does not.}
\label{tab:mnist}
\end{table*}

As shown in Figure~\ref{alg:training_procedure}, to train the model, we take a random training example from each class and combine them into a set of anchors. In each iteration, we select a batch of training examples and compute the distances between each anchor and the training examples in the batch. Then we construct the minibatch of triplets used to train the model. We only include triplets where the positive example is closer to the anchor than the negative example and the said negative example is within the $d$ closest to the anchor among negative examples that are farther from the anchor than the positive example. The variable $d$ is a hyperparameter and is known as the \emph{dynamic margin}, because the distance to the farthest negative example to the anchor that is included varies depending on the scale of the embedding. The reason we do not include triplets where the positive example is farther from the anchor than the negative example is because the model would tend to map all examples to zeros otherwise, which is in line with the observations of \citep{schroff2015facenet}. 

While training, we need to retrieve the closest examples to the anchor. To this end, we use an efficient $k$-nearest neighbour search algorithm that is scalable to large datasets and high dimensions known as Prioritized DCI~\citep{li2017fast}. 

\subsection{Embedding Normalization}

We normalize the output embedding of our model, so that the embedding vector for each example has an $\ell_2$ norm of $1$. This can be interpreted as making sure $k$-nearest neighbors returns the $k$ closest vectors by \emph{angle}, which in this case is the same as return the vectors that are closest by Euclidean distance. So, our classifier is essentially basing its decision on the maximum cosine similarity between the image embeddings. 

\section{Experiments}
\label{sec:experiments}

Most existing continual learning methods can be adapted to work in the \emph{class learning} setting, i.e.: each method is asked to classify among all possible classes across all tasks, but is given the identity of the task at training time. It is often non-trivial to adapt them for the \emph{discrete task-agnostic learning} or \emph{continuous task-agnostic learning} settings, where the task identity or even task boundaries are not given at training time. Even though the proposed method works in the continuous task-agnostic learning, to enable comparisons to a broad range of existing methods, we evaluate it in the more restrictive \emph{class learning} setting. 

More concretely, in the \emph{class learning} setting, training data from a subset of categories are presented to the learner in stages. The two-task setting consists of the following stages: in the first stage, a subset of the categories are presented (which will henceforth be referred to as Set A), and the learner can iterate over training data from these categories as many times as it likes. In the second stage, the remaining categories are presented (which will henceforth be known as Set B); while the learner can look at training data from these categories as often as it desires, it no longer has access to the training data from the categories presented in the first stage. The $n$-task setting can be defined analogously. We perform our evaluation in both the two-task and five-task settings, though for clarity we will describe the evaluation protocol in the two-task setting. 

Note that the setting under which evaluation is performed affects the reported performance of each method, and the same method can yield very different performance under different settings. In particular, some methods may work very well under more restrictive settings, but undergo a significant performance degradation under less restrictive settings. This is because different underlying strategies employed by various methods may be especially dependent on certain assumptions. For example, various existing methods work quite well under the \emph{task learning} setting when the task identity is given at test time and performance is reported for classification among categories within the particular task (and not among categories across all tasks). However, they exhibit a drastic performance drop under the \emph{class learning} setting when these assumptions are removed. Refer to~\citep{van2019three,hsu2018re} for an extensive study and discussion on changes in performance of various methods under changes to the learning setting. 

\subsection{Evaluation Protocol}

The training procedure for two-task \emph{class learning} is: 

\begin{enumerate}
    \item Train on Set A. 
    \item Starting from the parameters obtained from the first step, train on Set B. 
\end{enumerate}

We evaluate performance on Set A after each step in order to assess the extent each method retains its performance on Set A after training on Set B. 

\subsection{Datasets and Splits}

We evaluate the proposed method on three datasets, MNIST, CIFAR-10 and ImageNet. We consider five different datasets/splits: 

\paragraph{MNIST} 
We first consider MNIST, which is the simplest dataset. We divide it into 2 tasks: Set A consists of the odd digits and Set B includes the even digits.

\paragraph{CIFAR-10} 

\begin{table*}[t]
\centering
\footnotesize
\begin{tabular}{lccc}
\toprule 
Method & Set A: \textit{Original} & Set A: \textit{After Seeing Set B} & Set B\\
&&(Difference from \textit{Original})&\\
\midrule
  Our Method & 72.3\% & 62.3\%\ (-10.0\%) & 64.2\% \\
\midrule
\multicolumn{4}{l}{\emph{Forgetful \& Adaptive:}}\\
  Fine-tuning with dropout (ratio: 0.5) & 64.02\% & 0.0\%\ (-64.02\%) & 85.05\% \\
  Elastic weight consolidation (EWC) & 67.38\% & 0.0\%\ (-67.38\%) & 76.08\% \\
  Online EWC & 68.97\% & 0.0\%\ (-68.97\%) & 79.17\% \\
  Synaptic intelligence (SI) & 68.47\% & 0.0\%\ (-68.47\%) & 82.83\% \\
  Deep Generative Replay (DGR) & 69.13\% & 0.40\%\ (-68.73\%) & 83.75\% \\
  DGR with distillation & 68.22\% & 0.03\%\ (-68.19\%) & 85.00\% \\
  Replay-trough-Feedback (RtF) & 66.57\% & 0.20\%\ (-66.37\%) & 86.60\% \\
\midrule
\multicolumn{4}{l}{\emph{Not Forgetful \& Less Adaptive:}}\\
  Learning without Forgetting (LwF) & 69.28\% & 45.50\%\ (-23.78\%) & 55.53\% \\
\bottomrule
\end{tabular}
\caption{Performance of the proposed method compared to eight other continual learning methods on CIFAR-10. SGD with dropout, EWC, online EWC and SI suffer from catastrophic forgetting on Set A after seeing Set B, but were able to adapt well to the new task (Set B). DGR, DGR + distillation and RtF also forget on Set A, this could relate to the fact that CIFAR-10 data is more complex and it is difficult for these methods to train a generative model for replaying. LwF retains some knowledge learned on Set A; however, it does not adapt as well to Set B as our method.}
\label{tab:cifar}
\end{table*}

We then consider CIFAR-10 and take Set A to be the six animal categories (bird, cat, deer, dog, frog and horse), and Set B to be the four remaining categories corresponding to man-made objects (airplane, automobile, ship, truck). This split is designed to be more challenging than just a random split, since we don't expect many intermediate-level features to be shared across Sets A and B, making it easier to forget about Set A when training on Set B. 

\paragraph{CIFAR-10 (5 Tasks)} 

We also consider a variant of CIFAR-10 with 5 tasks, where each task corresponds to a pair of consecutive categories. This is more challenging than the two-task setting above, because many more examples would be seen after training on the earlier tasks, and so preventing forgetting on the earlier tasks is harder.

\paragraph{ImageNet (Random Split)}

Next we consider ImageNet, which is larger and more complex than CIFAR-10. We randomly select 100 classes to serve as Set A, and 100 classes to serve as Set B. 

\paragraph{ImageNet (Dog Split)}

Finally, we consider ImageNet and use a more challenging split. For Set A, we take it to be 120 categories that are not dogs; for Set B, we take it to be all 120 categories that correspond to particular breeds of dogs. Similar to the split we used for CIFAR-10, this split is designed to be more challenging by minimizing the semantic overlap between Sets A and B. 

\subsection{Network Architecture}

We use the same architecture for the proposed method and all baselines, except for the last layer. In the last layer the baselines use a fully connected layer followed by a softmax classification layer, whereas the proposed method uses a $k$-nearest neighbour classifier. For baselines that require the training of a deep generative model such as a variational autoencoder, we choose a symmetric architecture for the decoder, where the order of the layers in the encoder is reversed and convolutions are replaced with transposed convolutions. 

For each dataset, we use a network architecture that has proven to work well on the dataset. For MNIST, following \citep{van2019three}, we use a network with two hidden layers of 400 units each. We use this same architecture for our method and all baselines. For CIFAR-10, we use the architecture from the Tensorflow CIFAR-10 tutorial, which consists of two convolutional layers followed by three fully-connected layers. For ImageNet, we use the AlexNet architecture~\citep{krizhevsky2012imagenet}, which consists of five convolutional layers followed by three fully-connected layers. All networks are trained using Adam~\citep{kingma2014adam}. 

\begin{table*}[h]
\centering
\footnotesize
\begin{tabular}{lcccccc}
\toprule 
Method & Task 1 & Task 2 & Task 3 & Task 4 & Task 5 & Average\\
\midrule
 Our Method & 43.00\% & 16.20\% & 24.65\% & 48.85\% & 27.55\% & 32.05\% \\
\midrule
\multicolumn{7}{l}{\emph{Very Forgetful \& Adaptive:}}\\
 Fine-tuning with dropout (ratio: 0.5) & 0.0\% & 0.0\% & 0.0\% & 0.0\% & 92.85\% & 18.57\% \\
 Elastic weight consolidation (EWC) & 0.0\% & 0.0\% & 0.0\% & 0.0\% & 89.50\% & 17.90\% \\
 Online EWC & 0.0\% & 0.0\% & 0.0\% & 0.55\% & 88.30\% & 17.77\% \\
 Synaptic intelligence (SI) & 0.0\% & 0.0\% & 0.0\% & 0.0\% & 93.05\% & 18.61\% \\
\midrule
\multicolumn{2}{l}{\emph{Forgetful \& Adaptive:}}\\
 Deep Generative Replay (DGR) & 0.15\% & 1.05\% & 0.0\% & 0.10\% & 93.75\% & 19.01\% \\
 DGR with distillation & 0.05\% & 0.70\% & 0.0\% & 0.15\% & 93.15\% & 18.81\% \\
 Replay-trough-Feedback (RtF) & 0.0\% & 1.45\% & 0.85\% & 0.20\% & 94.10\% & 19.32\% \\
\midrule
\multicolumn{2}{l}{\emph{Not Forgetful \& Not Adaptive:}}\\
 Learning without Forgetting (LwF) & 93.55\% & 0.65\% & 0.0\% & 0.0\% & 0.0\% & 18.84\% \\
\bottomrule
\end{tabular}
\caption{Performance of the proposed method compared to eight other continual learning methods on 5-task CIFAR. SGD with dropout, EWC, online EWC and SI suffer from catastrophic forgetting on previous tasks after seeing newer tasks, but were able to adapt well to the last task. DGR, DGR + distillation and RtF forget more on the previous tasks compared to our method. LwF retains knowledge learned on first set; however, it fails to adapt to newer tasks.}
\label{tab:cifar5}
\end{table*}

\begin{table*}[h]
\centering
\footnotesize
\begin{tabular}{lcccc}
\toprule 
& \multicolumn{2}{c}{Our Method} & \multicolumn{2}{c}{Oracle Baseline}\\
\midrule
& \textit{Original} & \textit{After Seeing Set B} & \textit{Original} & \textit{After Seeing Set B}\\
&  & (Difference from \textit{Original}) &  & (Difference from \textit{Original})\\
\midrule
ImageNet (Random split) & 69.5\% & 53.5\% (-16.0\%) & 72.7\% & 30.2\% (-42.5\%) \\
ImageNet (Dog split) & 67.0\% & 44.7\% (-22.3\%) & 70.0\%& 30.4\% (-39.6\%) \\
\bottomrule
\end{tabular}
\caption{Performance comparison between our method and the vanilla baseline on ImageNet. Performance evaluated as top-1 percent correct on test data for the original task. Baseline represents the network trained with cross-entropy loss. ``Original'' denotes the network after being trained on the original task (classes in Set A) and ``After Seeing Set B'' denotes the network after it is fine-tuned on the new task (classes in Set B). See corresponding sections for more details on the dataSet And network used for each row.}
\label{tab:results}
\end{table*}

\begin{table*}[h]
\centering
\footnotesize
\begin{tabular}{lccc}
\toprule 
& Set A (\textit{Original}) & Set A (\textit{After Seeing Set B}) & Set B\\
\midrule
Original & 69.5\% & 53.5\% & 69.2\% \\ 
\midrule
Number of Positive Examples (5 $\rightarrow$ 30) & +0.4\% & +0.1\% & +0.3\% \\ 
Number of Negative Examples (40 $\rightarrow$ 10)& -0.9\% & -1.8\% & -0.6\% \\ 
Embedding Dimension (100 $\rightarrow$ 200)& -0.1\% & +0.1\% & -0.5\% \\
Embedding Dimension (100 $\rightarrow$ 50)& -1.3\% & -0.3\% & -0.1\% \\
\bottomrule
\end{tabular}
\caption{Test accuracy under the original hyperparameter setting and the change in test accuracy under different hyperparameter settings relative to the original setting. Increasing the number of positive examples (from 5 to 30) slightly improves performance. Decreasing the number of negative examples (from 40 to 10) slightly reduces test accuracy. Increasing the embedding dimension (from 100 to 200) results in similar performance as the original setting, whereas decreasing the embedding dimension (from 100 to 50) moderately lowers the performance. Overall, the method is fairly robust to hyperparameter changes.}
\label{tab:sensitivity}
\end{table*}

\begin{table*}[h]
\centering
\footnotesize
\begin{tabular}{lccc}
\toprule 
& Set A (\textit{Original}) & Set A (\textit{After Seeing Set B}) & Set B\\
\midrule
With All Components & 69.5\% & 53.5\% & 69.2\% \\ 
\midrule
No Normalization & -12.2\% & -2.4\% & -17.8\% \\ 
No Dynamic Margin & -9.5\% & +0.8\% & -14.8\% \\
\bottomrule
\end{tabular}
\caption{Test accuracy with all components, and changes in test accuracy relative to the original proposed method after various components are removed. Not normalizing the embedding output dramatically lowers test accuracy, and especially limits the ability to adapt to the new task (Set B). Disabling dynamic margin also greatly decreases the performance on Set B. Overall, we found that embedding normalization and dynamic margin are crucial for the method to be able to adapt well to new tasks.}
\label{tab:ablation}
\end{table*}

\subsection{Key Findings}

The performance of the proposed method and the baselines on various datasets and splits are shown in Tables~\ref{tab:mnist}, \ref{tab:cifar}, \ref{tab:cifar5} and \ref{tab:results}. The test accuracy on Set A before fine-tuning on Set B is reported in the sub-column titled ``Original''. We report the test accuracy after fine-tuning along with the change in accuracy after vs. before fine-tuning in the sub-column titled ``After Seeing Set B''. 

On MNIST and CIFAR-10, we compare to eight continual learning methods, including fine-tuning with dropout, Elastic Weight Consolidation (EWC), online EWC, Synaptic Intelligence (SI), Deep Generative Replay (DGR), DGR with Distillation, Replay-through-Feedback (RtF) and Learning without Forgetting (LwF). 

Table~\ref{tab:mnist} shows the results of the proposed method and existing continual learning methods on MNIST. MNIST is the dataset for which most existing continual learning methods are tuned, and so various baselines are expected to attain their peak performance on MNIST. Yet, as shown, regularization-based methods including EWC, online EWC and SI are very forgetful of the knowledge learned on the previous task (Set A) but can adapt to the new task (Set B). Replay-based methods (DGR, DGR + distillation, RtF) are adaptive as well, but are more forgetful compared to the proposed method. While these methods retain knowledge much better, they require training deep generative models that essentially learn the distribution of training data for the previous task. As a result, the performance of these methods depends on how difficult it is to train a good generative model of the data; while this is relatively easy on MNIST, it is much more difficult on CIFAR-10. Hence, performance of these methods is quite sensitive to the choice of dataset, as we will show in Table~\ref{tab:cifar}. Another replay method, LwF, retains knowledge from the previous task (Set A) well, however, it struggles with adapting to the new task (Set B). In contrast, the proposed method can both adapt to the new task and retain knowledge about the previous task, which is especially remarkable considering its simplicity and ability to operate without knowledge of task boundaries. 

Table~\ref{tab:cifar} show the results of the proposed method and existing continual learning methods on slightly more complex CIFAR-10 dataset. Regularization-based methods (EWC, online EWC, SI) and replay-based methods (DGR, DGR + distillation, RtF) all forget the knowledge from the previous task (Set A) but are adaptive to the new task (Set B). As mentioned above, these methods can no longer retain much knowledge because they depend on training deep generative models. LwF retains the knowledge on the first task (Set A) but it fails to adapt to the second task (Set B) compared to the proposed method. Table~\ref{tab:cifar5} shows the results of the proposed method and the same methods on a more challenging 5 tasks setting. Regularization-based methods (EWC, online EWC, SI) fail to retain knowledge from previous tasks and adapt to the last task. Replay-based methods (DGR, DGR + distillation, RtF) suffer more from catastrophic forgetting on previous tasks compared to the proposed method. LwF keeps the knowledge of the first task but is not adaptive to newer tasks.

We also evaluate our method on ImageNet. Because of the poor performance of the baselines on CIFAR-10, which is smaller and simpler than ImageNet, we compare to an \emph{oracle} that has access to Set A after training on Set B, and uses the data from Set A to fine-tune the last layer. Note that the oracle has an unfair advantage over the proposed method because it can fine-tune on Set A again after fine-tuning on Set B; despite this, the proposed method significantly outperforms the oracle as shown in Table~\ref{tab:results}. The proposed method was able to achieve $1.77\times$ the accuracy on Set A after seeing Set B compared to the baseline on the random split and improve the accuracy by $1.47\times$ on the dog split. The dog split is more challenging because Set A (non-dog images) and Set B (dog images) in that case are more semantically dissimilar. Since Set B only consists of dog images, fine-tuning on this data may wipe out much of the knowledge the model has learned about images in general. In contrast, since the random split does contain a random sampling of classes in each set, it is likely that fine-tuning on Set B preserves more general concepts about images that are still useful in classifying the images in Set A.

\subsection{Analysis}
We further examine the performance of the proposed method on the ImageNet dataset with random split.

\subsubsection{Sensitivity Analysis}
We varied the hyperparameter settings and took a look at the impact of changes in hyperparameters. We tried the following settings: reduce number of negative samples, increase number of positive sample and different embedding dimensions. As shown in Table~\ref{tab:sensitivity}, increasing the number of positive examples showed a slight improvement, whereas decreasing the number of negative examples resulted a slight performance drop. A similar trend can be observed with the embedding dimension, where proposed method does not benefit from the increase in embedding dimension, neither does it deteriorate significantly when the embedding dimension is decreased. Overall, we found the method to be reasonably robust to hyperparameter changes.

\subsubsection{Ablation Study}
We tested the proposed method by disabling some key components to determine their contributions. The following components are analyzed: embedding output normalization and dynamic margin. As shown in Table~\ref{tab:ablation}, by disabling embedding output normalization, the performance on both tasks drops significantly. When the dynamic margin is removed and replaced with a static margin, the method fails to capture new information from Set B. In summary, we found that embedding normalization and dynamic margin contribute significantly to the performance of the proposed method.

\section{Conclusion and Future Work}
\label{sec:conclusions}

As discussed in Section~\ref{sec:experiments}, we achieve consistent improvements over baselines across five different datasets and splits. These results suggest that the proposed method effectively addresses the catastrophic forgetting problem, and hint at a possible general strategy for reducing catastrophic forgetting, namely the idea of replacing the last layer of a neural net with a non-linear classifier. 

In the future, we would like to explore combining the proposed approach with other orthogonal approaches for tackling catastrophic forgetting. We also plan to explore applications that can benefit from continual learning approaches, such as reinforcement learning.

\bibliography{continual_learning}
\bibliographystyle{continual_learning}

\end{document}